# Corporate core values and social responsibility: What really matters to whom

Barchiesi, M. A., & Fronzetti Colladon, A.

# Corporate core values and social responsibility: What really matters to whom

Barchiesi, M. A., & Fronzetti Colladon, A.

**Abstract**

This study uses an innovative measure, the Semantic Brand Score, to assess the interest of stakeholders in different company core values. Among others, we focus on corporate social responsibility (CSR) core value statements, and on the attention they receive from five categories of stakeholders (customers, company communication teams, employees, associations and media). Combining big data methods and tools of Social Network Analysis and Text Mining, we analyzed about 58,000 Italian tweets and found that different stakeholders have different prevailing interests. CSR gets much less attention than expected. Core values related to customers and employees are in the foreground.

# 1. Introduction

The stakeholder theory is one of the main and most recurrently used approaches in social, environmental, and sustainability organization research (Frynas and Stephens, 2015; Silva et al., 2019); at the core of this approach there is the evolution from a corporate centric view - in which stakeholders are considered as subjects to be managed - to a network-based, relational and process-oriented perspective of company-stakeholder engagement (Andriof and Waddock, 2002; Steurer, 2006). According to the stakeholder theory there are two core questions that managers should ask themselves (Freeman, 1994): what is the purpose of the firm? What responsibility does management have to stakeholders? To answer the first question, managers need to itemize the shared sense of the value they generate, discovering what brings firm stakeholders together. The second answer leads managers to examine what kind of relationships they want and need with key stakeholders (Sundaram and Inkpen, 2004). In a certain sense, at the basis of stakeholder theory there is the idea of at least partially shared values and attitudes (Richter and Dow, 2017). However, each stakeholder chooses the level to which he/she provides or withholds sustenance to a brand or an organization (Niemann-Struweg, 2014). Company stakeholders can contribute to value creation in business (Dey et al., 2019; Hidalgo and Herrera, 2020; Pólvora et al., 2020); however, several studies proved that motivation to collaborate is heterogeneous across individuals and able to affect behaviors (Bridoux et al., 2011; Bridoux and Stoelhorst, 2014). Managers should periodically investigate the involvement of each category of stakeholder in organization core values – also with the purpose of being more effective in the application of their corporate social responsibility strategies. Indeed, at the basis of an integrated approach between Stakeholder Theory and CSR there is the assumption that companies act responsibly because of the pressure of their stakeholders (Delmas and Toffel, 2004; Fernandez-Feijoo et al., 2014; Huang et al., 2009; Shnayder et al., 2016) who have become significantly involved and demanding in corporate social responsibility practices (Barchiesi et al., 2018; Ranängen and Lindman, 2018; Silva et al., 2019). This study uses big data analytics to question this assumptions, and, in addition, it extends the possible applications of the Semantic Brand Score (SBS), a new measure of semantic importance (Fronzetti Colladon, 2018), which proves to be suitable for the analysis of core values and for the evaluation of stakeholders preferences.

Our purpose is to analyze the differences in stakeholders' interests with regard to company core values and corporate social responsibility specifically, by examining a particular place where company-stakeholder interactions take place, i.e. the social media Twitter. Combining methods and tools from text mining and social network analysis, we read the signals coming from social interactions and propose a new and easily replicable methodology to assess the importance of different core values, as they could differ in terms of stakeholder desirability.

This study evaluates the interest of stakeholders in different company core values – considering the Twitter platform – and specifically addresses the following research questions:

RQ1: Is Corporate Social Responsibility a stakeholders' prevalent interest?

RQ2: Are stakeholders more interested in CSR voluntary practices than in those prescribed by law?

Addressing these questions will provide evidence as whether corporate social responsibility, and especially the voluntary company contribution to the improvement of society and the environment, is really at the core of key stakeholders' interest. The underlying rationale is that CSR initiatives could be ineffective if they are not responsive to the different attitudes and interests of key stakeholders. Indeed, many authors recognize the necessity to engage with a range of stakeholders while planning and implementing CSR strategies (e.g., Lane and Devin, 2018; O'Riordan and Fairbrass, 2008; Pedersen, 2006). Stakeholders' involvement in CSR practices cannot be taken for granted. Each company has a specific view on what is its role in society, and expresses it in its core values. In order to be effective in CSR strategies, companies need to build a shared sense of their core values when this is missing, or adapt their actions to the values and norms of acceptable behavior for the social system they belong. Indeed, organization legitimacy depends on the congruence between the values implied by the company actions and those of the social system (Dowling and Pfeffer, 1975). In this paper, we analyze the perceptions of stakeholders about the organizational core values of the world's most admired companies, according to Fortune's 2013-2017 rankings. We present an analysis of data extracted from Twitter regarding five categories of stakeholders (customers, company communication teams tweeting on behalf of top management, employees, associations and media). The methodology we use could also be replicated for an almost real-time assessment of the importance of core values considering social media other than Twitter and other categories of stakeholders. In addition, a single firm could analyze stakeholders' perceptions about its own core values.

In order to categorize the core values of different organizations we referred to a classification of core values. A classification is an arrangement of entities into taxonomic classes or groups according to their common characteristics. The aim of a classification is, on the basis of precise criteria, to separate into diverse categories the entities used for the same purpose, following a logical order. For example, Agle and Caldwell (1999) classify values according to five levels of analysis (individual, organizational, institutional, societal and global); while Cording et al. (2014) refer to two dimensions of values, the espoused and the realized ones and to four combinations of these two dimensions (weak, under-promising, over promising and strong values). Depending on the research purpose and the logical order used, a classification might be more functional than others. Although in literature there are numerous classifications of core values (e.g., Agle and Caldwell, 1999; Cording et al., 2014; Hitlin and Piliavin, 2004), we chose to refer to the one described by Barchiesi and La Bella (2014), as it comprises six relevant stakeholders' orientations, and has a specific focus on corporate social responsibility, which is studied both in terms of citizenship and social responsibility. In this core value classification, core value orientations are the exemplifications of the nature of beliefs and thoughts that are subtended in core value expressions. For example, core value expressions such as "safety and health at work" and "a fun and rewarding place to work" may lead to different strategic perspectives, but share the same nature, i.e. the employee core value orientation. Consistently with Barchiesi and La Bella (2014), we refer to six core value orientations: customer/user, employee, economic and financial growth, excellence, citizenship (i.e. core values referring to the correct behavior that companies have to keep because of the shared norms of the systems where they operate), and social responsibility (i.e. values referring to companies' voluntary contribution to the improvement of society and environment).

The paper is organized as follows. In Section 2, we provide a theoretical background on stakeholder theory and its extension to corporate social responsibility. We also describe the classification of core values that we adopted for this study. In Section 3, we introduce the Semantic Brand Score and illustrate the approach we used to assess the importance of core values on social media. Our methodology is detailed in Section 4 and the results of our case study are presented in Section 5. The final section is devoted to discussing our findings and their implications.

## 2. Theoretical background

Stakeholders are groups or individuals that have legitimized interest in the actions and results of a firm and on whom the firm relies to realize its purposes (Freeman, 1984). According to stakeholder theory, successful firms are those able to create value for their stakeholders in a long term perspective (Freeman et al., 2010). Before developing corporate strategies, managers should actively explore their relationships with all stakeholders (Perrini and Tencati, 2006; Sprengel and Busch, 2011). The fundamental purpose of stakeholder management is to develop methods and practices to manage these relationships (Freeman and Mcvea, 2008). The literature on corporate social responsibility extends the scope of stakeholder analysis, placing business relationships side by side with societal and environmental relationships, because economic and noneconomic responsibilities are not distinct subjects, they are both part of corporate social responsibility (Carroll, 1979). According to the European Commission (*A renewed EU strategy 2011–2014 for*

*Corporate Social Responsibility*, 2011) a company is accountable for its impact on all relevant stakeholders, and the aim of CSR is maximizing the creation of value for all stakeholders and society. Thus, Corporate Social Responsibility can be integrated in stakeholder management theory to better understand the role (and relationships) of firms and organizations in society. However, considering each kind of stakeholder naturally (i.e. by their own natural character, without any external input) and equally (i.e. at the same level) interested in all corporate social responsibility practices could limit the effectiveness of an integrated approach between CSR and stakeholder theory (Madsen and Ulhøi, 2001). According to social psychologists and behavioral economists, individuals act according to their own "social value orientation", i.e. their preference about how to allocate resources between the self and the collectivity (Bridoux and Stoelhorst, 2014). Stakeholders' different backgrounds, representation schemes and purposes, lead them to different interpretations of the same situation and, consequently, to different preference systems (Ananda and Herath, 2003). While managers should balance all stakeholders' interests, and consider the needs and aspirations of all stakeholders in defining CSR strategies, it is also true that stakeholders have different bargaining power in influencing managers' choices (Bridoux and Vishwanathan, 2020). Indeed, a widely accepted normative basis of CSR is still missing (Campbell, 2007; Sarkar and Searcy, 2016) and companies choose their CSR initiatives according to their own aims and purposes (van Marrewijk, 2003), under the dynamic constraint to be legitimate in their social system (Dowling and Pfeffer, 1975). In summary, corporate social responsibility is about company core behavior, and company core behavior is about company core values. Company core values are concepts and beliefs that influence group behaviors. We can consider them as the most distinguishing characteristic of an institution (Williams, 2002), collective beliefs about what the entire enterprise stands for, takes pride in, and holds of intrinsic worth (Fitzgerald and Desjardins, 2004). Even if they are "central and enduring tenets of the organization" (Collins and Porras, 1994, p. 73), organizational core values - as well as the norms and values of the social system of reference - can change in terms of desirability over time and space (Gimeno et al., 2006; Parada et al., 2010;

Yoganathan et al., 2017). This change is induced by the pressure of competing interests amongst stakeholders (Urde, 2009; Yoganathan et al., 2017). Managers should accurately evaluate stakeholders' involvement in their organization's core values (Barchiesi and La Bella, 2014), as it could be hard to implement a strategy in contrast with those values that are the most important for the stakeholders (Pant and Lachman, 1998; Van Rekom et al., 2006).

### 2.1. Core Value Classification

In literature there are numerous taxonomies of core values (e.g., Agle and Caldwell, 1999; Cording et al., 2014; Hitlin and Piliavin, 2004). In our analysis, we use and extend the one theorized by Barchiesi and La Bella (2014), because it allows to compare different organizations with respect to diverse stakeholders' interests: what organizations do for customers/users (customers/user orientation); what organizations do for employees (employees orientation); what organizations do to obtain outstanding results (excellence orientation); what organizations do to obtain profits (economic and financial growth orientation); the integrity of an organization (citizenship orientation); company social responsibility (social responsibility orientation). Our distinction between citizenship and social responsibility is focused on the voluntariness of responsible actions, and not on the diverse typology of CSR practices, as for example social or environmental issues. Thereby it is possible to highlight an important aspect of CSR, emphasized by many authors (Piacentini et al., 2000; *Promoting a European Framework for Corporate Social Responsibilities*, 2001), i.e. the commitment to go beyond what is prescribed by the law.

In the following, we provide some examples of core values and their classification. Examples in the customer/user orientation are: "maintain reasonable prices" (J&J); "we are passionate about building strong, long-lasting client relationships. This dedication spurs us to go "above and beyond" on our client's behalf" (IBM); "Our clients' interests come first" (Goldman Sachs). An example of core value in the employees orientation is: "Leaders raise the performance bar with every hire and promotion. They recognize exceptional talent, and willingly move them throughout the organization. Leaders develop leaders and take seriously their role in coaching others" (Amazon). An example of core value in the excellence orientation is "We exhibit a strong will to win in the marketplace and in every aspect of our business" (Amex). An example of core values in the economic and financial growth orientation is: "We are committed to enhancing the long-term value of the investment dollars entrusted to us by our shareholders. By running the business profitably and responsibly, we expect our shareholders to be rewarded with superior returns. This commitment drives the management of our Corporation" (Exxon Mobil). Examples of core values in the citizenship orientation are: "We are honest, ethical, and trustworthy" (Microsoft); "We operate within the letter and spirit of the law" (Procter&Gamble); "We hold ourselves to uncompromising ethical and legal standards" (Marriott). Examples of core values in the social responsibility orientation are: "We want to help shape a better world and inspire people to live healthier lives" (Nestlè); "An important part of us is giving our time, talent, energy and resources to our community and society. Corporate community involvement is coordinated through Our Foundation. Our associate-led volunteer force uses the time and talents of associates to meet community needs through hands-on service" (Home Depot).

## 3. Assessing the Importance of Core Values on Social Media

Nowadays, a large variety of relationships with stakeholders takes place on social media (Benthaus et al., 2016; Bhimani et al., 2019; Mention et al., 2019; Misuraca et al., 2020; Risius and Beck, 2015; Tajudeen et al., 2018; Yang et al., 2020). Platforms such as Facebook or Twitter allow to identify and reach a wide range of stakeholders, engaging them in a real two-way interaction (Fujita et al., 2019; Manetti and Bellucci, 2016; Unerman and Bennett, 2004). In addition, these websites allow stakeholders' communication without the mediation of corporations (Rybalko and Seltzer, 2010), i.e. they can share their thoughts and ideas on many topics. For example, customers use social media to get information about the products they are willing to buy, to share their personal feelings about a brand, or to get post-sale assistance (Misuraca et al., 2020). Therefore, social media has become a repository of stakeholders' experiences, including opinions and perceptions about organizational core values, corporate sustainability practices and Sustainable Development Goals (Lee and Kim, 2021). Twitter, in particular, showed to be an elective communication channel of organizations with their stakeholders and of stakeholders with each other (Mamic and Almaraz, 2013; Rybalko and Seltzer, 2010). For example, when an organization states that one of its principles of business operations is "water" (see for example Nestlé), the discourse on Twitter about the sustainable use of water and the continuous improvement in water management could inform about the stakeholders' level of engagement for that particular organizational core value.

In order to assess the importance of a discussion topic on social media – for some category of stakeholders – one possibility is to consider the frequency of use of the terms which identify that topic. Similarly, popularity of a page on Facebook can be measured by looking at its number of likes, or counting the comments to its posts (De Vries et al., 2012). However, limiting the analysis to the count of frequencies can be too restrictive.

Fronzetti Colladon (2018) has recently proposed a new approach for the measurement of *brand importance*, which is a construct that can be used to capture the importance of a word (brand), or set of words (concept/topic), in a discourse. The indicator the author offers is the Semantic Brand Score (SBS), which has three components: prevalence, diversity and connectivity. The frequency of use of the words representing a specific topic is measured by prevalence, which however is just one of the dimensions of the SBS. Mentioning a core-value-related topic in a high number of tweets increases its visibility and the probability that other readers will recall it. It is a signal of the importance attributed by the writer to the topic and of her/his familiarity with it, which is related to the concept of brand awareness (Fronzetti Colladon, 2018; Keller, 1993). However, it may happen that a concept is frequently mentioned but always in association with the same set of words. Diversity is the second component of the SBS and measures the heterogeneity of the words surrounding each discourse topic. The higher the number of different words used in the context of a specific core value orientation (resulting in the discourse being less repetitive and probably more informative), the higher its diversity. This measure is consistent with previous research showing the positive effect of a higher number of associations on brand strength (Grohs et al., 2016). Lastly, connectivity expresses the "embeddedness" of a core value orientation in the discourse of a set of stakeholders. Connectivity reflects the ability of a brand/concept to link different parts of a discourse and keep them together. It could be considered as the 'brokerage' power of a specific concept (Fronzetti Colladon, 2018). Indeed, a core value could either be peripheral or at the core of a discourse, regardless of its prevalence and diversity scores. Connectivity was similarly used in past research with the intent of assessing brand popularity (Gloor et al., 2009).

According to the SBS, an orientation (i.e. the set of core values belonging to it) is highly important when it keeps the different parts of a discourse together and it is frequently mentioned and surrounded by heterogeneous textual associations. Prevalence, diversity and connectivity are all necessary to capture the importance of a core value orientation (Fronzetti Colladon, 2018).

The use we make of the Semantic Brand Score is new: to the extent of our knowledge, this metric was not yet applied to assess the importance of company core values. The SBS was specifically conceived to evaluate of the importance of brands, or broader concepts, through the analysis of (big) textual data. This well suits the case of core value orientations (Fronzetti Colladon, 2018), which are broad concepts representable by sets of keywords instead of single words. As described in Section 4, the calculation of the SBS needs to transform texts into networks, where each node is a word. In the case of broad concepts, multiple nodes can be merged to evaluate their joint importance.

## 4. Methodology

### 4.1. Data Collection and Classification

The microblogging platform Twitter has drawn the attention of many scholars, as it has the potential to capture the dialogue of a large set of users, who are active or potential stakeholders. Indeed, many studies showed that Twitter is one of the major communication channels of companies with stakeholders (Mamic and Almaraz, 2013; Rybalko and Seltzer, 2010). We crawled Twitter for a period of two months, in order to extract the discourse about the core values of the world's most admired companies, according to the magazine Fortune. We only considered those companies which were in the ranking for five consecutive years, from 2013 to 2017; they are listed in Table 1.

Annually Fortune, together with the Korn Ferry Institute, surveys about 15,000 executives, directors, and security analysts to identify and rank the World's Most Admired Companies, rated on nine attributes: innovation, people management, use of corporate asset, social responsibility, quality of management, financial soundness, long-term investment, quality of products/services, and global competitiveness. The candidate companies must have $10 billion in revenue and rank among the largest by revenue within their industry. The companies that were in the ranking for 5 consecutive years all had high revenues and a strong reputation. The Fortune's index was criticized by some authors (Mahon, 2002), because of its main focus on financial performance from the viewpoint of CEO's and industrial analysts. This group is nevertheless one of the most interesting set of companies to analyze with respect to core values and stakeholders' engagement. They have clearly stated core values, are well-known, well-respected and involve a significant number of stakeholders.

Obviously, companies are free to express their core values as they prefer, without any categorical reference. This did not make it easy to attribute a core value expressed through long sentences to a single orientation, but the taxonomy we used was able to capture the different orientations of core values without overlaps, in the vast majority of cases. In the very few cases of core values referring to more than one orientation, we worked by concepts. As an example,

Facebook's core value statement "We believe that a more open world is a better world because people with more information can make better decisions and have a greater impact. That goes for running our company as well. We work hard to make sure everyone at Facebook has access to as much information as possible about every part of the company so they can make the best decisions and have the greatest impact" refers both to social responsibility (for the first part) and employees (for the second part). Concepts are "to allow everyone in the world to have access as much information as possible" and "a workplace in which everyone has access to as much information as possible".

Successively, we picked up the keywords that better conveyed the meaning expressed by each statement, considering also their synonyms. This helped us build the Twitter search queries and organize word clusters for the SBS analysis. In the following, we provide just some examples of keywords for each orientation:

- Employees: "teamwork", "work freely", "learning environment", "accountability", "diversity", "inclusion", "career", "fair treatment";
- Customers: "product quality", "service quality", "product safety", "technological innovation", "responsiveness", "customer satisfaction";
- Excellence: "success", "market leadership", "excellence", "continuous improvement", "best practices", "highest standards";
- Economic and Financial Growth: "long-term value", "profitability", "shareholders", "stockholders", "superior returns", "capital investment opportunities";
- Citizenship: "integrity", "legal compliance", "ethics", "transparency", "obey the law", "honesty";
- Social Responsibility: "information access", "environmental responsibility", "social responsibility", "water use", "healthier lives", "care access", "natural resource protection", "better education".

These keywords were used to create complex search queries, combined with other terms related to the business world – such as "company", "business" or "employees" – in order to exclude tweets that could be out of topic. Following this approach, we collected about 58,000 tweets, written in Italian and posted by more than 26,000 users. In the observation period, we did not detect any event that could significantly bias the data collection – such as a scandal or an announcement of a revolutionary product. With the help of two independent annotators, we analyzed Twitter profiles to classify users into different stakeholder categories. The classification was carried out reading the description of user profiles – where it was indicated, for example, if the account was of an employee, a newspaper or a company. When this information was not available, the annotators read the user tweets. The annotators first worked independently and then met to discuss their judgments and find an agreement. Five different categories of stakeholders emerged from this analysis: customers or prospective customers, employees, company communication teams (i.e. employees who tweeted from the official Twitter account of their company, speaking on behalf of the top management), media (such as journalists, information websites or newspapers), associations (such as consumers' associations, environmental associations or unions). No other significant category of stakeholders was found and the number of tweets from other sources was negligible. Indeed 0.4% of tweets were excluded as these were spam (mostly produced by bots).

| Name | Industry |
|---|---|
| Accenture | Information Technology Services |
| Alphabet (Google) | Internet Services and Retailing |
| Amazon.com | Internet Services and Retailing |
| American Express | Consumer Credit Card and Related Services |
| Apple | Computers |
| Berkshire Hathaway | Insurance: Property and Casualty |
| BMW | Motor Vehicles |
| Boeing | Aerospace and Defence |
| Caterpillar | Construction and Farm Machinery |
| Coca-Cola | Beverages |
| Costco | General Merchandisers |
| Disney | Entertainment |
| Exxon Mobil | Petroleum Refining |
| Facebook | Internet Services and Retailing |
| FedEx | Delivery |
| General Electric | Industrial Machinery |
| Goldman Sachs Group | Megabanks |
| Home Depot | Specialty Retailers |
| IBM | Information Technology Services |
| Intel | Semiconductors |
| Johnson & Johnson | Pharmaceuticals |
| JPMorgan Chase | Megabanks |
| Marriott International | Hotels, Casinos, and Resorts |
| Microsoft | Computer Software |
| Nestle | Consumer Food Products |
| Nike | Apparel |
| Nordstrom | General Merchandisers |
| PepsiCo | Consumer Food Products |
| Procter & Gamble | Soaps and Cosmetics |
| Southwest Airlines | Airlines |
| Starbucks | Food Services |
| Target | General Merchandisers |
| Toyota Motor | Motor Vehicles |
| Unilever | Soaps and Cosmetics |
| UPS | Delivery |
| Walmart | General Merchandisers |
| Whole Foods Market | Food and Drug Stores |

**Table 1** – World's Most Admired Companies from 2013 to 2017 (Fortune Magazine ranking)

### 4.2. Core Value Importance

The calculation of the Semantic Brand Score is based on the construction of a network of co-occurring words (Fronzetti Colladon, 2018). We briefly recall here the main steps of the process.

Using the SBS BI web app[1] (Fronzetti Colladon and Grippa, 2020), tweets were preprocessed to remove stop-words and punctuation, and convert every word to lowercase. We also identified bigrams and trigrams. Word affixes were removed using a snowball stemming algorithm developed for the Italian language (Porter, 2006; Willett, 2006). Lemmatization is an alternative approach to stemming, which can be useful with morphologically rich languages like Italian (Kettunen et al., 2005; Misuraca and Spano, 2020), but requires an adequate vocabulary.

Subsequently, the collection of text documents (tweets of a group of stakeholders) was transformed into a network graph, where each node is a word, linked to a second word, if the two co-occurred in the same tweet and within a range of 7 words (Fronzetti Colladon, 2018). Arcs were weighted considering the co-occurrence frequency of each pair of nodes. For example, if the phrase "Aurora is beautiful" is repeated five times, there will be an arc connecting the (stemmed) words 'aurora' and 'beauti', whose weight will be 5. Final graphs were pruned removing those arcs with a weight lower than 2, as probably representing random or unimportant co-occurrences. Words representing each core value orientation were merged into single nodes. The full network, comprising all the tweets, was also analyzed. Trying a different co-occurrence range did not change our results significantly.

We calculated prevalence as the frequency of use of the terms representing each core value orientation. For diversity and connectivity we worked on the above mentioned graphs, and used metrics of social network analysis (Wasserman and Faust, 1994). In particular, diversity was calculated as the *distinctiveness centrality* of the orientation nodes, i.e. counting their number of links to the other words in the discourse and the uniqueness of these links (Fronzetti Colladon and Naldi, 2020). The higher this number, the more heterogeneous the textual associations with the orientation analyzed. For connectivity, we referred to the metric of *betweenness centrality* (Freeman, 1979), also considering the inverse of arc weights to express network distances.

---

[1] Available at: https://bi.semanticbrandscore.com

Accordingly, network paths took into account the frequency of co-occurrence of each pair of words and connectivity was calculated as the weighted betweenness centrality of orientation nodes, using the algorithm proposed by Brandes (2001). This metric counts how many times a core value orientation lies in the paths that connect the other nodes. It can be intended as the 'brokerage power' of an orientation, measuring its ability to link different parts of the discourse (Fronzetti Colladon, 2018).

SBS was finally obtained from the sum of the standardized scores of prevalence, diversity and connectivity – attributing the same importance to each of the three dimension, as in the original presentation of the metric (Fronzetti Colladon, 2018). Each score has been standardized by subtracting the mean and dividing by the standard deviation.

All the operations of text pre-processing, network construction and SBS calculation were carried out using the SBS BI web app (Fronzetti Colladon and Grippa, 2020).

### 4.3. Language Sentiment

To complement the information coming from the SBS, we calculated a sentiment variable, to evaluate the positivity or negativity of the language used by each category of stakeholder, with regard to each core value orientation. Sentiment may vary in the range [-1,1], where 0 represents a neutral tweet, 1 a totally positive one and -1 a tweet conveying very negative feelings. Sentiment has been calculated thanks to the software Condor (Gloor, 2017), which uses a machine learning algorithm trained on a large set of tweets (Brönnimann, 2014, 2013; Gloor, 2017).

## 5. Results

Table 2 shows preliminary statistics for our sample: 55.5% of the tweets were produced by customers or prospective customers. Official company accounts (communication teams) are the second most active tweeters, followed by employees. Media – such as online newspapers – produced 9.5% of the tweets, whereas consumers' associations and environmental associations contributed for 4.6% of the tweets. Spam and negligible tweets from other sources represented only 0.4% of our sample and were excluded from the analysis. Spammers were identified following the procedure suggested in the work of Fronzetti Colladon and Gloor (2019): they send a very high number of tweets, sometimes all through the 24 h, but they are almost never mentioned in the tweets of others. Spammers usually follow a much larger number of accounts than the number of their followers and produce message of low quality content (that our annotators classified as spam). Language sentiment was positive on average and higher in media messages. The fact that average sentiment is often close to zero does not mean that we did not find any polarized tweet (more positive or negative). However, positive messages balanced and outnumbered negative ones. In addition, we observed that, in a good number of tweets, opinions were expressed in a neutral way. Some more sentiment variation can be observed in Table 3, where the indicator is calculated in association with each core value orientation.

|  | Volume of tweets | Average Length (words) M (SD) | Average Sentiment M (SD) |
| --- | --- | --- | --- |
| **Associations** | 4.6% | 16.86 (3.77) | 0.115 (0.244) |
| **Communication Teams** | 17.1% | 16.47 (3.69) | 0.105 (0.212) |
| **Customers/Prospective Customers** | 55.5% | 17.99 (4.18) | 0.112 (0.240) |
| **Employees** | 12.9% | 17.16 (3.88) | 0.112 (0.224) |
| **Media** | 9.5% | 15.77 (3.87) | 0.150 (0.256) |
| **Bot (spam)/Others** | 0.4% | 18.03 (3.92) | NA |

*Note*. M = mean; SD = standard deviation.

**Table 2.** Sample statistics

Table 3 presents the main results of our study, i.e. the relative importance scores of core value orientations for each category of stakeholder. Overall, among core values orientations, customers had the largest importance. Customers also seem to pay a large attention to employee policies (those expressed by core values) and to the economic and financial growth of the companies they are interested in. Communication teams, on the other hand, communicate mostly about the core values belonging to the customers orientation (relative SBS is 58.86%) and the ones belonging to employees orientation – which is however less than half important (20.13%).

Answering the question "what really matters to whom?", we could say that the customers core value orientation is the top priority for every category of stakeholders, except for media which pay more attention to companies' economic and financial growth. Another point of general attention is the one regarding the employees orientation. Values of excellence, on the other hand, were always the bottom priority for each stakeholder. Social responsibility got much less attention than expected: this orientation is ranked $5^{th}$ out of 6 orientations, in the overall discourse. Those more concerned with social responsibility were, not surprisingly, the associations. Customers, on the other hand, seem to value citizenship more than social responsibility – i.e. they care about the correct behavior of companies, but they are probably less demanding regarding companies' voluntary contribution to the improvement of environment and society. Results for the separate components of the SBS – prevalence, diversity and connectivity – were mostly consistent with their combined score. The sentiment, as already mentioned, was always positive on average, except for the excellence orientation in the discourse of associations, maybe due to the associations' role, i.e. protection of the collective interests. Excellence was the orientation with the lowest sentiment on average, even if still positive. Associations were critical about employees' rights and therefore had a lower sentiment in the employees orientation. Lastly, the advertising of social responsibility practices (those expressed by core values) conveyed more positive feelings than the discourse about firms' alignment with societal rules (citizenship).

| Measure | Core Value Orientation | Associations | Communication Teams | Customers | Employees | Media | Overall |
|---|---|---|---|---|---|---|---|
| Semantic Brand Score | Employees | 28.11% | 20.13% | 30.28% | 26.28% | 26.46% | 27.57% |
| | Customers | 43.19% | 58.86% | 35.06% | 46.80% | 31.21% | 39.70% |
| | Excellence | 1.76% | 3.19% | 3.92% | 2.94% | 1.90% | 3.65% |
| | Eco&Fin Growth | 11.61% | 11.06% | 18.46% | 12.09% | 32.14% | 17.80% |
| | Citizenship | 5.55% | 2.91% | 8.34% | 7.00% | 4.09% | 6.84% |
| | Social Responsibility | 9.78% | 3.86% | 3.94% | 4.88% | 4.21% | 4.45% |
| Prevalence | Employees | 29.29% | 16.69% | 32.07% | 23.45% | 22.80% | 27.44% |
| | Customers | 38.70% | 59.73% | 30.70% | 46.45% | 26.46% | 37.29% |
| | Excellence | 2.48% | 3.34% | 4.06% | 3.26% | 2.00% | 3.56% |
| | Eco&Fin Growth | 12.84% | 13.56% | 20.40% | 14.57% | 40.79% | 20.33% |
| | Citizenship | 5.70% | 2.80% | 8.57% | 6.86% | 3.80% | 6.74% |
| | Social Responsibility | 10.98% | 3.86% | 4.21% | 5.40% | 4.16% | 4.64% |
| Diversity | Employees | 26.71% | 23.80% | 28.27% | 28.31% | 28.61% | 27.37% |
| | Customers | 40.49% | 48.68% | 31.92% | 37.76% | 32.44% | 33.64% |
| | Excellence | 2.48% | 5.10% | 6.18% | 4.62% | 3.45% | 6.22% |
| | Eco&Fin Growth | 12.29% | 11.42% | 16.38% | 12.34% | 22.27% | 15.68% |
| | Citizenship | 8.01% | 5.27% | 11.31% | 10.48% | 6.68% | 10.34% |
| | Social Responsibility | 10.02% | 5.72% | 5.94% | 6.49% | 6.55% | 6.76% |
| Connectivity | Employees | 28.04% | 20.81% | 30.11% | 27.72% | 28.92% | 27.93% |
| | Customers | 51.69% | 68.87% | 43.39% | 56.91% | 35.98% | 48.66% |
| | Excellence | 0.11% | 0.86% | 1.52% | 0.72% | 0.19% | 1.24% |
| | Eco&Fin Growth | 9.33% | 7.20% | 18.17% | 8.68% | 31.20% | 16.67% |
| | Citizenship | 2.84% | 0.46% | 5.15% | 3.46% | 1.84% | 3.55% |
| | Social Responsibility | 7.99% | 1.79% | 1.67% | 2.49% | 1.88% | 1.95% |
| Sentiment | Employees | 0.09 | 0.21 | 0.13 | 0.19 | 0.12 | 0.14 |
| | Customers | 0.17 | 0.17 | 0.14 | 0.22 | 0.15 | 0.16 |
| | Excellence | -0.08 | 0.20 | 0.08 | 0.08 | 0.24 | 0.11 |
| | Eco&Fin Growth | 0.20 | 0.21 | 0.18 | 0.16 | 0.22 | 0.19 |
| | Citizenship | 0.13 | 0.19 | 0.14 | 0.14 | 0.18 | 0.15 |
| | Social Responsibility | 0.21 | 0.17 | 0.18 | 0.17 | 0.21 | 0.18 |

**Table 3.** Importance of Core Value Orientations

## 6. Discussion and Conclusions

The application of stakeholder theory to sustainability management is based on the assumption that companies act responsibly because of the pressure of their stakeholders (Delmas and Toffel, 2004; Fernandez-Feijoo et al., 2014; Huang et al., 2009; Shnayder et al., 2016), who have become significantly involved and demanding in corporate social responsibility practices (Barchiesi et al., 2018; Ranängen and Lindman, 2018; Silva et al., 2019). According to this assumption, companies' key stakeholders should be naturally interested to environmental and social issues. In this paper, we question this statement, investigating the differences in stakeholders' interests with regard to companies' core values and in particular to corporate social responsibility.

We used a novel indicator, the Semantic Brand Score (Fronzetti Colladon, 2018). This metric combines methods and tools of Semantic and Social Network Analysis, and it is suitable for the analysis of big textual data. We looked at about 58,000 tweets posted by five different categories of stakeholders (customers, employees, communication teams, media and associations) during two months.

Even though the sentiment was positive on average, the interest of stakeholders was distributed unequally across the six orientations (employees, customers, excellence, economic and financial growth, citizenship and social responsibility). The relatively high number of tweets was a first signal of the general interest of stakeholders in core-value-related topics. Surprisingly, the most active tweeters were not the companies' communication teams, but customers – with tweets three times more numerous. Combined with the positive sentiment, this seems to suggest that the analyzed core values were not empty principles for stakeholders. In general, stakeholders' attention was mainly focused on core values regarding customers, followed by those about employees and economic and financial growth (see Figure 1 summarizing our findings).

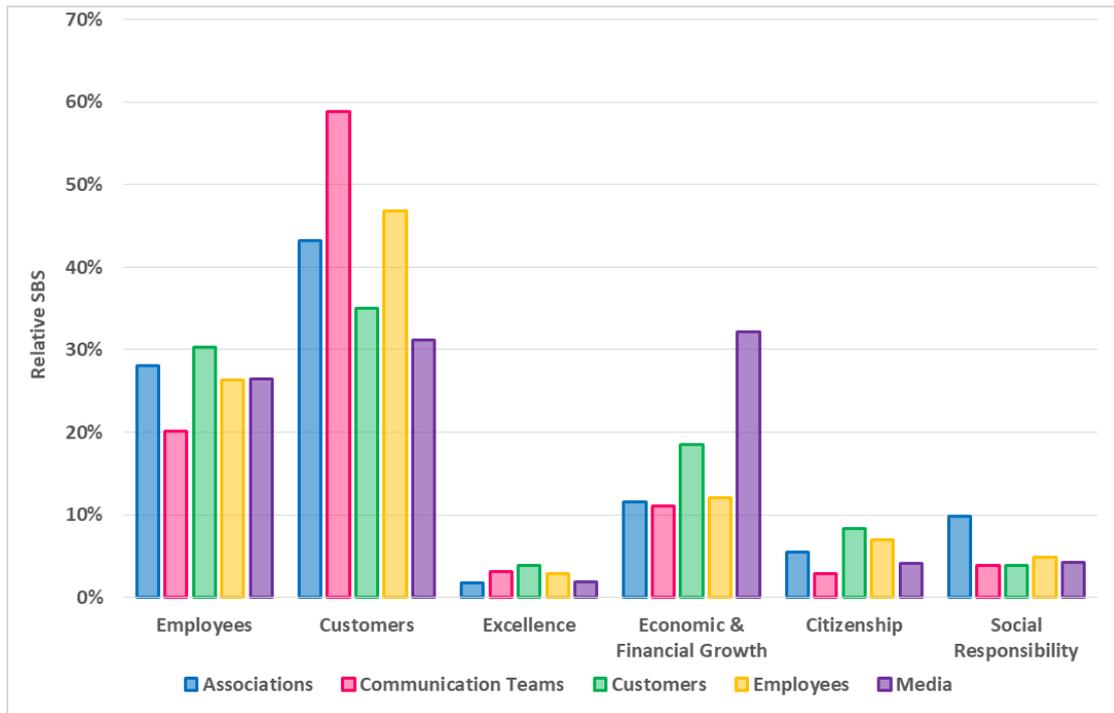

**Figure 1.** SBS of core value orientations

Different stakeholders' categories showed different prevailing interests. Surprisingly, the importance attributed by stakeholders to CSR (citizenship and social responsibility together) was low; only core values related to excellence got less attention. This answers, at least for this case study, to our first research question (RQ1). CSR was not a stakeholders' prevalent interest. This is aligned with studies that analyzed the Twitter discourse with different techniques (Barchiesi and Fronzetti Colladon, 2021) and poses the interesting question whether corporate social responsibility is really at the core of stakeholders' interests. Our findings suggest that the link between CSR initiatives and the proactive interest of stakeholders is far from being automatic, especially when these initiatives are not supported by a pervasive communication plan. This result is particularly interesting because the analyzed companies durably maintain a strong reputation, also due to their corporate social responsibility practices. The relationship between stakeholder engagement and corporate responsibility is complex (Greenwood, 2007) and it could be an error to take for granted the interest and constructive attitude of stakeholders towards corporate social responsibility actions.

Our methodology allows to detect and measure the difference in stakeholders' interest towards voluntary or prescribed-by-law social responsibility actions (RQ2). In other words, in terms of stakeholders' engagement, is it more important for a company "to be good" or "not to be bad"?

Despite the great emphasis given to the voluntariness aspect of CSR in literature (e.g., Piacentini et al., 2000; *Promoting a European Framework for Corporate Social Responsibilities*, 2001), customers and employees seem more engaged with company law compliance. By contrast, associations and communication teams put more emphasis on social responsibility, but they produce a significantly inferior number of tweets than customers and employees (22% vs 67%).

In order to be effective in CSR strategies, responsible companies should leverage on sustainability interests shared with stakeholders. This fundamental aspect is often neglected, or at least not sufficiently emphasized by companies. According to many authors (e.g., Schoeneborn et al., 2020), CSR communication has a constitutive role in creating, maintaining, and transforming CSR practice. Indeed, listening to stakeholders' voices on a particular matter, not only allows to confront multiple viewpoints on what should be done about that matter, but also allows new elements to be revealed (Cooren, 2020). With a strategic "bottom-up" process, activated by interactions, companies have the possibility to modify the definition of social legitimacy, in order to increase convergence with their own values (Lee et al., 2018).

In our research, we observe that company communication teams were the stakeholders with the lowest SBS scores for citizenship and social responsibility. It could be not enough to announce and to correctly run a CSR practice; it may also be necessary to work to improve stakeholders' engagement and stimulate their interest. Therefore, a timely and continuous assessment of core value rankings, made possible by our methodology, has important implications for the formulation and implementation of CSR and business strategies.

This work has some limitations which could be addressed in future research: a broader number of tweets written in languages different from Italian could be collected and analyzed. It could be that values important for Italian-speaking stakeholders differ from those relevant for people of other countries or cultures. The calculation of the SBS can easily be adapted to languages other than Italian. In addition, the importance of core values and corporate social responsibility could be evaluated looking at textual data extracted from other media sources – e.g. online newspapers, TV transcripts, Facebook and other social media platforms. It is important to notice that our study is devoted to the assessment of core values importance and offers a new methodology for their ranking. However, speculating on the reasons behind stakeholders' interests is out of scope and certainly deserves new dedicated research.

In this study, we calculated language sentiment through the software Condor (Gloor, 2017) which uses an algorithm specifically trained for tweets (Brönnimann, 2014, 2013; Gloor, 2017). However, other approaches are possible and could be tested in future research to profile stakeholders and enrich the view on their opinions and on the positivity or negativity of their expressions (e.g., Greco and Polli, 2020).

The use of Twitter, as well as other media, could introduce some biases, due to the fact that only a subset of people use Twitter. In addition, some employees might not be allowed to tweet on their labor conditions. Similarly, customers of B2B firms might be underrepresented on this media, if compared with customers of B2C companies. There might be differences between sectors. Accordingly, future research could extend our study, considering other media and/or communication channels with stakeholders. We cannot generalize our findings to other communication channels, media, companies or business settings, without carrying out further analyses.

In our analysis of Twitter, we found 5 categories of stakeholders (customers, employees, communication teams, media and associations) and no tweets from other categories. With our findings we do not intend to neglect that other categories of stakeholders exist (Freeman and Mcvea, 2008); it could just be that the categories we did not find make a limited usage of Twitter, or were not interested in the topics of our analysis. This could be an interesting area of exploration for future research. In any case, our methodology is not limited to the analysis of a specific set of stakeholders. The SBS can be used to evaluate core value importance for potentially all stakeholders, with the only requirement of them producing texts or communications transformable into textual data - such as the transcripts of face-to-face interviews. Moreover, while the analysis of core values of the world's most admired companies has its advantages, results might change if considering different sets of companies. Future studies could test this possibility and see if stakeholders' interests are differently oriented. Lastly, the analysis could be repeated considering a finer core value classification – for example, the Social Responsibility dimension could be furtherly split into people and environment orientations.

Our work looks at stakeholder management with new lenses and adds to the research on the applications of the Semantic Brand Score and on Semantic and Social Network Analysis of big data. In this study, the SBS has been used in a new context – to analyze broader concepts and not just single words representing a brand – and it has proved to be effective for the assessment of stakeholders' interest toward companies' core values.

**Acknowledgements**

The authors are grateful to Claudia Colladon for her help in revising this manuscript. The authors are grateful to Valeria Battarelli and Benedetta Pinzuti for their help in the data collection process. Andrea Fronzetti Colladon gratefully acknowledges financial support from the University of Perugia under Grant "Fondo Ricerca di Base 2020", project n. RICBA20LT ("Strumenti di analisi e gestione a supporto delle Smart Companies nell'era dell'Industria 4.0").